\documentclass[runningheads]{llncs}
\usepackage{graphicx}
\usepackage{amsmath,amssymb} 
\usepackage{color}
\usepackage{multirow}
\usepackage{enumitem}
\usepackage{booktabs}
\usepackage{subcaption}
\usepackage{makecell}

\begin{document}
\pagestyle{headings}
\mainmatter

\def\ACCV20SubNumber{479}  
\makeatletter
\renewcommand*{\@fnsymbol}[1]{\ifcase#1\or*\else\@arabic{\numexpr#1-1\relax}\fi}
\makeatother
\title{Learning to Adapt to Unseen Abnormal Activities under Weak Supervision} 
\titlerunning{Learning to Adapt to Unseen Abnormal Activities}
%
\author{Jaeyoo Park\thanks{These authors contributed equally.} \and 
Junha Kim\textsuperscript{*} \and 
Bohyung Han}
\authorrunning{J. Park, J. Kim, and B. Han}
%
\institute{ECE \& ASRI, Seoul National University, Korea \\
\email{\{bellos1203,junha.kim,bhhan\}@snu.ac.kr}}

\maketitle


\begin{abstract}
We present a meta-learning framework for weakly supervised anomaly detection in videos, where the detector learns to adapt to unseen types of abnormal activities effectively when only video-level annotations of binary labels are available.
Our work is motivated by the fact that existing methods suffer from poor generalization to diverse unseen examples.
We claim that an anomaly detector equipped with a meta-learning scheme alleviates the limitation by leading the model to an initialization point for better optimization.
We evaluate the performance of our framework on two challenging datasets, UCF-Crime and ShanghaiTech. The experimental results demonstrate that our algorithm boosts the capability to localize unseen abnormal events in a weakly supervised setting.
Besides the technical contributions, we perform the annotation of missing labels in the UCF-Crime dataset and make our task evaluated effectively.


\end{abstract}


\section{Introduction}
\label{sec:introduction}

\begin{figure}[t]
	\centering
	\includegraphics[width=\linewidth]{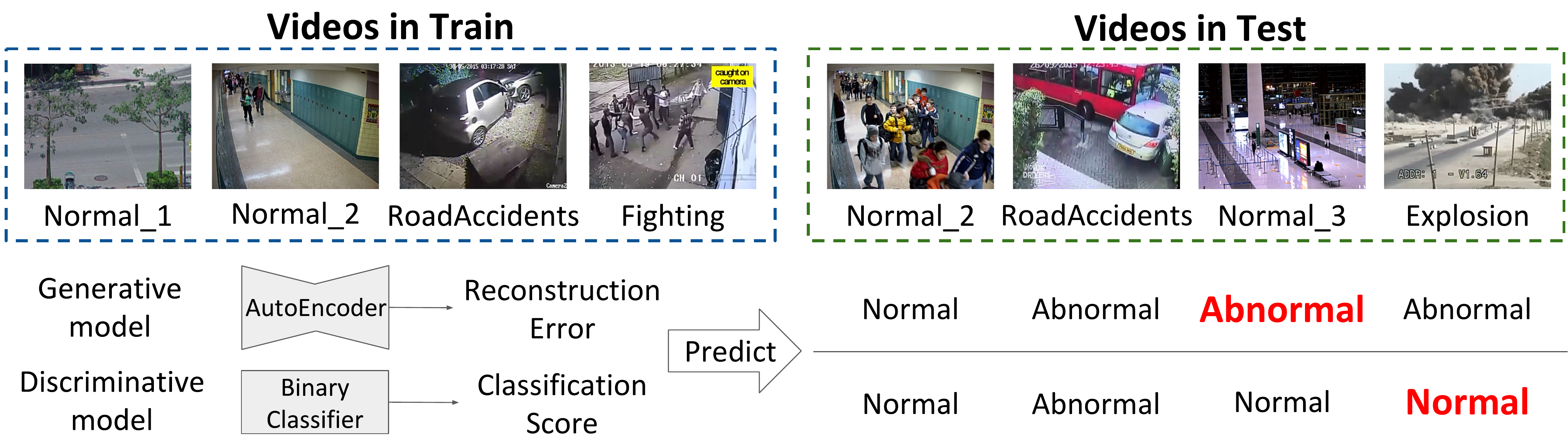}
	\caption{
		Limitation of the existing abnormal event detection approaches in videos.
		The generative models ({\it e.g.}, auto-encoder) attempt to learn normal patterns in training data; they successfully reconstruct the normal videos seen during training (Normal\_2) while they fail to reconstruct the videos captured from new viewpoints (Normal\_3).
		Meanwhile, the discriminative approaches ({\it e.g.}, binary classifier) focus on classifying each frame into two classes, abnormal and normal, by learning abnormal patterns from the given data.
		Therefore, the discriminative classifiers detect the abnormal events that have been seen during training (RoadAccidents) while they fail to recognize unseen types of abnormal events (Explosion) at test time.
		The bold-faced letters in red correspond to the wrong predictions of individual approaches.
	}
	\label{fig:concept}
\end{figure}

Humans easily identify unusual events from a video by generalizing prior knowledge spontaneously despite the ill-defined nature of anomaly detection.
On the contrary, computer vision algorithms rely on an extensive learning process based on a large number of annotated training examples to obtain a model for abnormal event detection.
There exist various approaches proposed for anomaly detection in videos.
The methods based on generative models~\cite{gong2019memorizing,hasan2016learning} claim the capability to reconstruct normal patterns while \cite{sultani2018real,zhong2019graph} propose discriminative techniques based on binary classifiers.
Despite the significant advance in anomaly detection on videos~\cite{gong2019memorizing,hasan2016learning,sultani2018real,zhong2019graph,antic2011video}, existing methods in both categories may suffer from critical drawbacks.
A recent study~\cite{sultani2018real} presents that generative approaches are not suitable for the recognition problems on videos with substantial scene variations since they are prone to predict unseen normal patterns as abnormal. 
Also, the generated videos often have limited diversity, especially having the same viewpoint as the cameras used to construct training examples.
On the other hand, the discriminative classifiers may not be robust to unseen types of normal or abnormal activities.
In particular, they can detect the predefined types of abnormal events only and tend to overfit to training data. 
Fig.~\ref{fig:concept} illustrates the limitations of the existing methods mentioned above.

Since it is infeasible to collect the videos containing all kinds of normal and abnormal activity patterns, the detector should be able to spot the eccentricity even with limited prior information.
Here, one crucial question arises.
When we learn a model to detect unseen patterns in videos, how can we take advantage of prior knowledge?
The simplest solution would be pretraining a model using the data of seen patterns followed by fine-tuning it with the examples of unseen types.
To address this problem in spite of a practical limitation---weak diversity of training examples, we formulate anomaly detection as learning to adapt to various unseen abnormalities rather than learning the universal representation.
To this end, we harness the meta-learning concept~\cite{finn2017model,raghu2019rapid}, which claims that the model parameters of a deep neural network can be located at a desirable initial point for better optimization, not necessarily fast convergence, by simulating the learning process for adaptation to new data distribution.
By constructing learning episodes containing diverse abnormal events, where the variance across individual examples is large, the model learns to reach the appropriate initialization point, which leads the model to adapt well to novel abnormal events. 



Moreover, we explore whether detecting unseen abnormal events can be effectively performed under weak supervision in the meta-learning framework.
Since it is expensive to obtain precise annotations of temporal locations of individual abnormal events in videos, we prefer constructing base-learner models using the examples with video-level binary labels, normal vs. abnormal.
Note that we aim to localize abnormal activities in the temporal domain via learning a model based on binary annotations of abnormality in video level.


We validate the proposed training scheme on two challenging datasets, UCF-Crime~\cite{sultani2018real} and ShanghaiTech~\cite{liu2018ano_pred}.
Since the UCF-Crime dataset provides temporal annotations of abnormal events only for a small portion of videos, we annotated the examples without ground-truths for our experiments. 
The experimental results show that the proposed algorithm outperforms the baseline in detecting novel abnormal events.
The source codes and new annotations are available at our project page\footnote{\url{https://cv.snu.ac.kr/research/Learning-to-Adapt-to-Unseen-Abnormal-Activities/}}.

%
The main contributions of the proposed approach are summarized below:
\begin{itemize}[label=$\bullet$]
	\item
	We formulate anomaly detection in videos as a learning-to-adapt task to unseen abnormal activities to address the limitations of existing approaches. \\

	\item 
	We propose a novel meta-learning approach under weak supervision, where the base-learner utilizes video-level binary labels only for training, while the final model estimates the localization information of unseen abnormal events. \\
	
	\item
	We labeled the missing ground-truths for temporal locations of abnormal events in the UCF-Crime dataset. \\
	
	\item
	The experiment on UCF-Crime with label augmentations and ShanghaiTech shows that our method is effective in learning novel types of abnormal events. 
\end{itemize}

The rest of the paper is organized as follows.
We first discuss related work in Section~\ref{sec:related}.
The overall procedure and the experimental results with their analysis are described in Section~\ref{sec:Method} and \ref{sec:experiments}, respectively.
Section~\ref{sec:conclusion} concludes this paper.



\section{Related Work}
\label{sec:related}

\subsection{Anomaly Detection}
\label{sub:anomaly}
Many researchers have been interested in anomaly detection in the video~\cite{gong2019memorizing,hasan2016learning,sultani2018real,zhong2019graph,antic2011video,kratz2009anomaly,wu2010chaotic,zhao2011online}. 
Given a video, the detector localizes unexpected incidents that are rarely observed. 
The task is challenging due to its ill-defined nature, its innate complexity, and the diversity of examples. 
 
The advances in generative modeling techniques based on deep neural networks allow us to construct the anomaly detector in a generative manner~\cite{hasan2016learning,lu2013abnormal,xu2015learning,zhao2017spatio}.
They attempt to find the general pattern of in-distribution data points with the generative models such as auto-encoder~\cite{gong2019memorizing,hasan2016learning,nguyen2019anomaly} and generative adversarial network~\cite{vu2019robust}.
Based on the assumption that abnormal events are rare, the generative models learn how to reconstruct normal and usual patterns. 
These models consider the examples that have large reconstruction errors as the out-of-distribution samples. 
However, they assume that all the videos have the same viewpoint. 
As a result, the models are prone to overfit to training data and predict unaccustomed normal patterns as abnormal.

A recent study~\cite{sultani2018real} claims that the classic generative approaches are unable to generalize normal patterns captured by the camera from a novel viewpoint. 
It also introduces a novel dataset for anomaly detection, UCF-Crime, which consists of more complex and diverse events than existing ones.
Based on the dataset, \cite{sultani2018real,zhong2019graph} suggest predicting abnormality scores in a discriminative manner. 
They treat the anomaly detection task as a binary classification problem under weak supervision, where the model classifies whether the video contains abnormal or normal events based only on video-level labels. 
Specifically, \cite{sultani2018real} proposes a binary classifier based on multiple instance learning, and \cite{zhong2019graph} employs a label noise cleaner using a graph convolutional neural network.
Nevertheless, those methods still suffer from a lack of generalizability, especially when they face unseen types of abnormality.
This fact raises the need for a reasonable initial model that handles unseen abnormality effectively.
Hence, we propose a meta-learning framework to obtain basic information from prior knowledge. 


\subsection{Meta-Learning} 
The objective of meta-learning is to realize the learn-to-learn capability, where the meta-learner supervises the learning process of the base-learner~\cite{finn2017model,lake2017building,thrun1998learning}.
The common approaches to address this problem include 1) metric learning-based methods, where the meta-learner focuses on the similarity metric within the task~\cite{snell2017prototypical,sung2018learning,vinyals2016matching}, 2) memory augmented methods, where the meta-learner stores training examples or class embedding features~\cite{mishra2017simple,munkhdalai2017meta,oreshkin2018tadam,santoro2016meta}, and optimization-based methods, where the meta-learner is directly parameterized by the information from the base-learner (e.g., gradients, etc,.)~\cite{finn2017model,andrychowicz2016learning,antoniou2018train,rusu2018meta}. 
Our work employs a popular optimization-based framework, Model-Agnostic Meta-Learning (MAML)~\cite{finn2017model}.
Further description of MAML will be provided in Section~\ref{sec:Meta-training}.
Following the recent works in other applications that take advantage of meta-learning schemes~\cite{choi2017deep,gui2018few,park2018meta,shaban2017one,wang2019panet,yan19meta}, we facilitate anomaly detection in videos using the meta-learning framework. 
Our work is distinct from existing models because we construct the meta-learning model upon base-learner under weak supervision, \textit{i.e.,} when only video-level labels are available.
There is a prior study that focuses on the generalization capability of meta-learning in domain generalization task~\cite{li2018learning}.
Recently, \cite{lu2020few} addresses an anomaly detection problem based on meta-learning.
However, its direction is different from ours in the sense that it attempts to learn the normality of scenes using generative models, while our approach aims adapt to novel anomaly via discriminative models.


\section{Method}
\label{sec:Method}

\subsection{Overview}
\label{sec:Overview}
Our goal is to learn an anomaly detection model that adapts to novel types of abnormal events effectively using weakly labeled examples.
Since it is infeasible to capture common abnormal patterns generally acceptable in the videos with huge diversity, we formulate anomaly detection as learning to adapt to various abnormal events rather than learning universal representations. 

We assume that there exists a video dataset $\mathcal{D}_{base}$ with two different types of annotations per video---binary label whether a video contains abnormal events or not, and $c_{base}$ abnormal event categories (subclasses) for the positive videos.
We learn to identify the initial model parameters optimized for adapting to novel abnormal events in the videos that belong to $\mathcal{D}_{novel}$ with $c_{novel}$ subclasses.
Note that the subclasses in $\mathcal{D}_{base}$ and $\mathcal{D}_{novel}$ are disjoint although we do not use the subclass information in the training procedure of the proposed framework. 
%
Another assumption is that, given a video $v=\{v_i\}_{i=1}^{N} \in \mathcal{D}~(=\mathcal{D}_{base}\cup\mathcal{D}_{novel})$ with $N$ segments, we only have the video-level label $y \in \{0,1\}$, where 1 indicates that the video has at least one abnormal segment, 0 otherwise.
Note that our weakly-supervised detector should predict per-segment label, $\hat{y}=\{\hat{y}_i\}_{i=1}^{N}$, without segment-level or localization ground-truths.   

We propose to harness meta-learning to boost the localization accuracy of unseen abnormal events based on weak supervision. 
We claim that simple knowledge transfer from pretrained models may not work well in our scenario since the variations of abnormality is significant and the prior knowledge obtained from seen abnormal events is difficult to be generalized to unseen ones.
Hence, by exploiting meta-learning based on the episodes with large variations, we alleviate the limitation of transfer learning and facilitate to learn models for unseen anomaly detection via meta-testing.
Specifically, we construct an episode by sampling a small subset of videos from $\mathcal{D}_{base}$, and perform an iteration of meta-training using the episode.
In the meta-testing phase, we fine-tune the model using the videos sampled from $\mathcal{D}_{novel}$ to obtain the final model.
Note that the entire training procedure relies only on video-level binary class labels.

The rest of this section describes the details about the individual components of our framework, which include 1) the base anomaly detector relying on weak labels, and 2) the meta-learning algorithm to obtain better generalizable models.

\subsection{Weakly Supervised Anomaly Detector}
\label{sec:Weakly Supervised Anomaly Detector}

We adopt the anomaly detection method proposed in~\cite{sultani2018real} as our base detector. 
The detector learns to score how abnormal each video segment is under weak supervision. 
The score of each segment is given by a binary classifier distinguishing between abnormal and normal events.
To train a segment-wise anomaly detector based only on video-level annotations, we employ Multiple Instance Learning (MIL) with a ranking loss.

\subsubsection{MIL}
The concept of MIL is employed in our problem to learn the rank between normal and abnormal bags.
We divide a video into $N$ segments, each of which is denoted by $v_i$ ($i = 1, \dots, N$).
A video $v=\{v_i\}_{i=1}^{N}$ with $N$ segments is regarded as a positive bag  $\mathcal{B}_a$ if at least one of the segments is abnormal, {\it i.e.}, $\exists i, y_i=1$.  
Otherwise, the video is normal and its segments construct a negative bag $\mathcal{B}_n$.
The segments in $\mathcal{B}_a$ and $\mathcal{B}_n$ pass through a scoring function $f(\cdot)$, which consists of three fully-connected layers with ReLUs and sigmoid functions, to predict abnormality scores.

\subsubsection{Ranking loss}
We employ a ranking loss for MIL as in \cite{sultani2018real}, which produces higher scores for abnormal segments than normal ones. 
Since segment-level labels are not available in our setting, the loss for a pair of a positive bag and a negative one is defined by the segments with the maximum scores in both bags as
\begin{equation}\label{eq:1}
\mathcal{L}(\theta ; \{ \mathcal{B}_a,\mathcal{B}_n \})= \max(0, m-\max_{v_i \in \mathcal{B}_a} f(v_i ; \theta) +\max_{v_i \in \mathcal{B}_n} f(v_i ; \theta)),
\end{equation}
where $\theta$ denotes model parameters, $v_i$ means the $i$-th segment in a bag, and $m$ indicates the score margin between the two bags. 
In addition, the loss function has two regularization terms---a temporal smoothness loss and a sparsity loss.
The former encourages temporally adjacent segments to have similar scores, while the latter enforces only a small subset of segments in a video to have high scores upon the assumption that abnormal activities rarely happen in videos.
By combining all the loss terms, the final loss function is given by
\begin{align}\label{eq:total_loss}
\mathcal{L}(\theta ; \{ \mathcal{B}_a,\mathcal{B}_n \}) &= \max(0, m-\max_{v_i \in \mathcal{B}_a} f(v_i ; \theta) +\max_{v_i \in \mathcal{B}_n} f(v_i ; \theta)) \nonumber\\
&+\lambda_1 {\sum_{v_i \in \mathcal{B}_a}(f(v_i ; \theta)-f(v_{i+1} ; \theta)) ^2} +\lambda_2 {\sum_{v_i \in \mathcal{B}_a} {f(v_i ; \theta)}},
\end{align}
where $\lambda_1$ and $\lambda_2$ are the hyperparmaters to control the impact of individual terms.
Following \cite{sultani2018real}, we set $m=1$ and $\lambda_1 = \lambda_2 = 8 \times 10^{-5}$ throughout our training procedure.
Note that, since $f(v_i ; \theta)$ is the output of a sigmoid function and always positive, the last term in Eq.~\eqref{eq:total_loss} is equivalent to $\ell_1$ norm of a segment-wise score vector.

\subsubsection{Training base detector}
We train the anomaly detector based on the objective function in Eq.~\eqref{eq:total_loss} using $\mathcal{D}_{base}$.
To train the detector, we split a video into multiple segments, where each segment consists of 16 consecutive frames, and extract 3D convolutional features from I3D networks~\cite{carreira2017quo} pretrained on the Kinetics dataset.
We represent each variable-length video using 32 non-overlapping transformed features as described in \cite{sultani2018real} and feed them to our base detector model for training.


\begin{figure}[t]
	\centering
	\includegraphics[width=\linewidth]{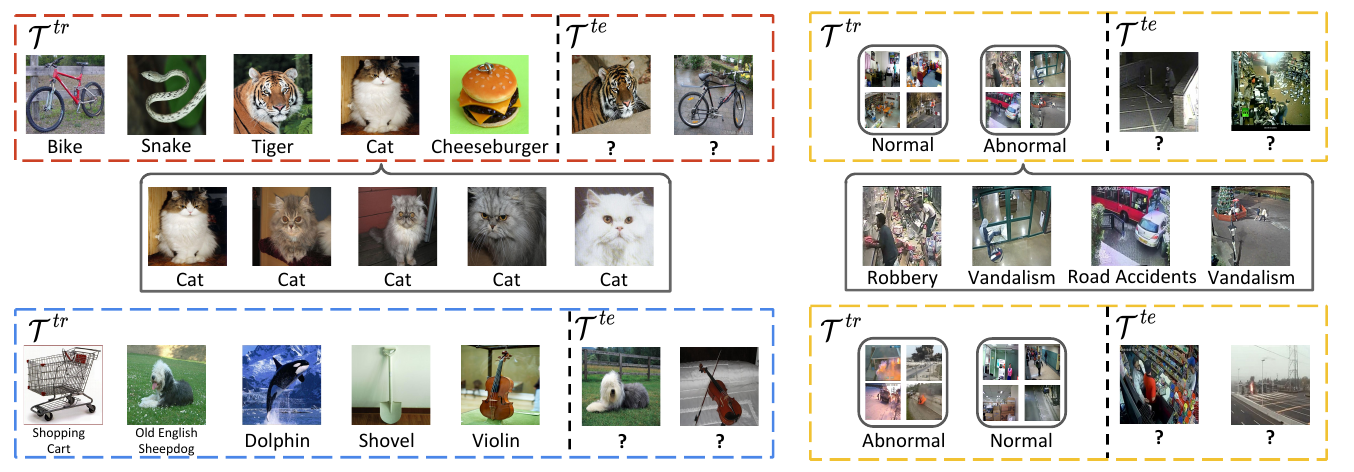}
	\centerline{\small \quad (a) Conventional episodes \quad \quad \quad \quad \quad \quad \quad \quad \quad (b) Our episodes}
	\caption{
		Comparison of the methods to construct episodes between the conventional $N$-way $K$-shot classification and our anomaly detection.
		(a) The tasks are different across episodes and the intra-class variation is relatively small.
		The images are sampled from \textit{miniImageNet}~\cite{ravi2016optimization} dataset.
		(b) Every episode is for binary classification between abnormal and normal classes.
		The abnormal class consists of the subclasses in $\mathcal{D}_{base}$, so the intra-class variation is large.
	}
	\label{fig:Episode}
\end{figure}

\subsection{Meta-training}\label{sec:Meta-training}
In the meta-training phase, our goal is to make the model learn to adapt to novel types of abnormal examples by repeatedly simulating the learning procedure using the data sampled from the distribution with large intra-class variations.
To achieve this goal, we adopt a meta-learning approach based on MAML~\cite{finn2017model}.
Our meta-learning scheme aims to find an optimal set of initial model parameters, which is suitable for adapting to unseen types of data.
Since there is no external meta-learner in MAML, the model parameters are solely updated by the gradient descent method.

\subsubsection{Episode in anomaly detection}\label{Episode}
We first describe how to construct episodes for anomaly detection. 
Most of the few-shot classification studies formulate an episode as a classification task, where a model for each episode is optimized for a unique set of classes, and have the target tasks for meta-testing separate from the ones for meta-training.
We refer to this kind of strategy as the conventional meta-learning in the rest of this section.

In contrast, all the tasks in anomaly detection are identical: binary classification between \textit{normal} and \textit{abnormal}.
We sample both normal and abnormal videos from $\mathcal{D}_{base}$ with ${c}_{base}$ subclasses to construct a task $\mathcal{T}$.
Note that the sampled abnormal videos belong to any subclass but the subclasses for meta-training and meta-testing should be disjoint.
The major difference between the conventional and our meta-learning lies in the source of diversity.
The intra-class variation in anomaly detection is much larger than that of the conventional meta-learning.
While the existing few-shot learning studies attempt to generalize models over the task distribution, our approach focuses on the generalization of the model over the data distribution within the individual classes, abnormal and normal, during meta-training. 
The difference between the conventional meta-learning and our approach is illustrated in Fig.~\ref{fig:Episode}.

\subsubsection{Training method}
We construct each task $\mathcal{T}_{i}$, which is divided into training and testing denoted respectively by $\mathcal{T}_{i}^{trn}$ and $\mathcal{T}_{i}^{tst}$, by sampling abnormal and normal videos from $\mathcal{D}_{base}$.
Using the training and testing splits, meta-training is performed by the bi-level optimization on the base detector.

We first, as in a typical training scenario, adapt the base detector to $\mathcal{T}_{i}^{trn}$ based on the objective function in Eq.~\eqref{eq:total_loss}. 
This adaptation step is referred to as the \emph{inner loop}. 
For the $i$-th task $\mathcal{T}_i$, the model parameters of the base detector, denoted by $\theta$, are then updated to $\Tilde{\theta}_{\mathcal{T}_i}$ by the gradient descent method using the loss function $\mathcal{L}_{\mathcal{T}}(\theta ; \mathcal{T}_{i}^{trn})$, which is expressed as
\begin{equation}\label{eq:3}
\Tilde{\theta}_{\mathcal{T}_i} = \theta-\alpha\nabla_\theta\mathcal{L}_{\mathcal{T}_i}(\theta ; \mathcal{T}_{i}^{trn}),
\end{equation}
where $\alpha$ is the learning rate for the base detector.

Next, the adapted base detector is evaluated by $\mathcal{T}_{i}^{tst}$, and the meta-learner is optimized using the resulting error.
Since the meta-optimization contains the adaptation step, it is also referred to as the \emph{outer loop}.
The meta-learner is optimized by $\mathcal{L}_{\mathcal{T}}(\tilde{\theta}_\mathcal{T} ; \mathcal{T}^{tst})$.
In MAML-based approaches, the meta-update is performed by updating the model parameters, denoted by $\theta$, of the base detector before the adaptation step using the meta-objective function, which is given by
\begin{align}\label{eq:4}
\min_\theta \sum_{\mathcal{T}_i \sim p(\mathcal{T})}  \mathcal{L}_{\mathcal{T}_i}(
\tilde{\theta}_{\mathcal{T}_i} ; \mathcal{T}^{tst}_{i})
= \sum_{\mathcal{T}_i \sim p(\mathcal{T})}  \mathcal{L}_{\mathcal{T}_i}(
\theta - \alpha \nabla_\theta \mathcal{L}_{\mathcal{T}_i}(\theta ; \mathcal{T}_{i}^{trn}) ; 
\mathcal{T}^{tst}_{i}).
\end{align}
Therefore, the model parameters are meta-updated as

\begin{equation}\label{eq:5}
\theta = \theta-\beta\nabla_\theta\sum_{\mathcal{T}_i \sim p(\mathcal{T})} \mathcal{L}_{\mathcal{T}_i}(
\tilde{\theta}_{\mathcal{T}_i} ;
\mathcal{T}^{tst}_{i}),\\   
\end{equation}
where $\beta$ is the meta-learning rate.

\subsection{Meta-testing}
In the meta-testing stage, we evaluate whether the model adapts to the novel types of abnormal events well. 
To this end, we fine-tune the model for the abnormal events in ${c}_{novel}$ subclasses, which are disjoint from $c_{base}$ normal subclasses, by constructing episodes for meta-testing using sampled examples from $\mathcal{D}_{novel}$.
Since we do not have a validation set $\mathcal{D}_{novel}^{val}$ due to the small size of the datasets, we perform 10-fold cross-validation by exploiting $\mathcal{D}_{novel}^{trn}$ to decide the number of iterations for fine-tuning.


\newcolumntype{M}[1]{>{\centering\arraybackslash}m{#1}}

\section{Experiments}
\label{sec:experiments}

\subsection{Datasets}
\label{sec:Datasets}
We conduct the experiments on two benchmark datasets, UCF-Crime~\cite{sultani2018real} with our label augmentation and ShanghaiTech~\cite{liu2018ano_pred}.

\subsubsection{UCF-Crime} 
This large-scale dataset consists of real-world surveillance videos captured in various circumstances.
It contains 13 subclasses of abnormal events including \textit{Abuse}, \textit{Arrest}, \textit{Arson}, \textit{Assault}, \textit{Burglary}, \textit{Explosion}, \textit{Fighting}, \textit{RoadAccidents}, \textit{Robbery}, \textit{Shooting}, \textit{Shoplifting}, \textit{Stealing}, and \textit{Vandalism}. 
The dataset has 1,900 untrimmed videos, including 950 abnormal videos and 950 normal ones.

There exist a couple of critical limitations in this dataset that hamper direct compatibility with our task.
First, the subclass distribution in the original training and testing splits given by \cite{sultani2018real} for anomaly detection is severely imbalanced.
Hence, we conduct our experiments using the action recognition split provided by \cite{sultani2018real}.
In the action recognition split, every subclass has 38 videos for training and 12 videos for testing.
Second, \cite{sultani2018real} provides the temporal durations of abnormal events for the test videos in its anomaly detection split while some videos in the test set of the action recognition split do not have such annotations.
To make the dataset complete for performance evaluation, we annotate the ground-truth intervals of abnormal events for some videos in the dataset.

\subsubsection{ShanghaiTech} 
This is a medium-scale dataset composed of 437 videos from 13 different scenes. 
Since all training videos are normal, we use a new split proposed by \cite{zhong2019graph}.
In addition, we employ this dataset only for meta-test since there are not a sufficient number of videos containing abnormal events for meta-training.
We believe that the experiment in this dataset shows the cross-dataset generalization performance of the proposed method.

\begin{table}[t]
\caption{Over-estimated performance issue in the existing evaluation method. The AUC score of our base detector is approximately 84\% when evaluating the entire test videos using the anomaly split in \cite{sultani2018real}.
The value is slightly higher than the one reported in \cite{sultani2018real} because we employ two-stream features from I3D network~\cite{carreira2017quo} for video representations.
However, when we exclude the normal videos from the test set, the performance drops to about 68\%.}
\label{table: overestimation}
\begin{center}
\scalebox{0.95}{
\renewcommand{\arraystretch}{1.1}
\setlength\tabcolsep{10pt}
\begin{tabular}{c|c|c}
	\toprule
	Class          & AUC (\%) & \# of test samples \\
	\hline
	Abnormal        & 68.35   & 140                    \\
	Abnormal+Normal & 84.39   & 290   \\
	\bottomrule                
\end{tabular}
}\vspace{-5mm}
\end{center}
\end{table}


\subsection{Evaluation Metric and Protocol}
\label{sec:Evaluation Metric}
Following the previous works~\cite{hasan2016learning,sultani2018real,zhong2019graph}, we draw the frame-wise receiver operation characteristic (ROC) curve and compute its area under curve (AUC) score.
However, our evaluation method is different from the existing ones in the following two parts.

First, we only evaluate the AUC performance on the abnormal videos.
Since there is a significantly larger number of normal frames than abnormal ones, especially if we count both abnormal and normal videos, performance evaluation using the videos in both classes leads to a biased result towards accuracy over-estimation as illustrated in Table~\ref{table: overestimation}.
Therefore, we exclude normal videos for the computation of the AUC scores in our experiments.
Note that we use the original splits and annotations instead of the revised ones to obtain the statistics.

Second, we evaluate the average frame-wise AUC score for each video while existing methods estimate the scores using all frames collected from all videos in their test datasets.
This is because the overall performance is often dominated by a small subset of extremely long videos, which are as long as $10^5$ frames and substantially longer than the average length, about $4 \times 10^3$ frames.

\subsection{Experimental Settings and Implementation Details}
\label{sec:Experimnetal Settings and Implementation Details}
To validate our claim that meta-learning provides a proper initialization point, we compare the following three scenarios, which are given by fine-tuning the detector on $\mathcal{D}_{novel}$ starting from 1) the randomly initialized model, 2) the pretrained model on $\mathcal{D}_{base}$, and 3) the meta-trained model on $\mathcal{D}_{base}$.

For the experiments, we re-implemented the detector proposed in \cite{sultani2018real} and use it as the base learner.
Our implementation is identical to \cite{sultani2018real} except the following three parts.
First, we utilized the pretrained two-stream I3D features~\cite{carreira2017quo} trained on the Kinetics dataset instead of C3D features~\cite{tran2015learning} employed in \cite{sultani2018real}; the optical flows are computed by the TVL1 algorithm~\cite{zach2007duality} and the fusion of two modalities---RGB and optical flow---is given by the concatenation of their features.
Second, we removed the dropout layers~\cite{srivastava2014dropout} since training the MAML model~\cite{finn2017model} was unstable.
Finally, we used the Adam optimizer instead of Adagrad.

For pretraining, we sampled 30 videos from both the abnormal and normal classes to form a mini-batch.
After splitting $\mathcal{D}_{base}$ into train and validation videos following the action recognition split, we trained the model with a learning rate $10^{-3}$ until the validation AUC score arrives at the peak.

For meta-training, we construct each episode using 10 samples for training and 30 for testing from both categories, abnormal and normal classes.
The learning rate of the inner loop is set to $10^{-3}$ while the learning rate for the outer loop, which is a meta-learning rate, is set to $10^{-5}$.
We trained the model for 3,000 outer iterations with meta-batch size 15, and used an SGD optimizer for inner loop optimization.

Fine-tuning is performed on $\mathcal{D}_{novel}$ regardless of the initialization methods with the learning rate $10^{-3}$.
We fine-tuned the model for 300 iterations at maximum and performed 10-fold cross-validation to choose the best model.

\begin{table*}[t!]
	\caption{
		AUC score (\%) comparisons among three different scenarios on each target subclass.  
		The fine-tuning process of all the compared methods is identical while the initial point of fine-tuning is different. 
		In the algorithm denoted by S, the model is fine-tuned from a random scratch model.
		In the scenario P, the model is pretrained with $\mathcal{D}_{base}$ before fine-tuning.
		Two versions of the meta-learning approach, denoted by M$_S$ and M$_G$, which performs meta-training with $\mathcal{D}_{base}$ to obtain the initial model, correspond to two different model selection strategies, ``sampling'' and ``global''.
		Details of ``sampling'' and ``global'' are described in Section~\ref{sec:Quantitative Results}.
		The bold-faced numbers correspond to the best accuracy for each subclass.
	}
	\label{table:result}
\begin{center}
	\scalebox{0.82}{
	\hspace{-0.3cm}
		\begin{tabular}{@{}c|c|ccccccccccccc|c@{}}
			\toprule
		 	Split &  Algo. & Abus & Arre & Arso & Assa& Burg & Expl & Figh & Road & Robb & Shoo & Shop & Stea & Vand & Avg \\ \midrule
			\multirow{4}{*}{1}     & S   &    62.99   &    67.91    &    56.93   &    80.05     &     72.02     &     63.62      &     70.94     &        73.19       &    77.86     &    75.81      &       57.70      &     66.77     &      72.86     &    69.13     \\
			& P  &    69.71   &    67.57    &  60.22 & 81.18    &    77.51     & 71.85     &     70.65      &     \textbf{77.11}     &        80.48       &     82.69    &     52.37     &      65.41       &      74.61    &     71.64               \\
			& M$_S$   &   \textbf{70.93}   &    \textbf{72.05}    &   \textbf{61.26}    &  \textbf{82.67}       & \textbf{81.08}         &    \textbf{73.32}       &   \textbf{71.35}       &     76.72          &   \textbf{82.56}      &     \textbf{82.85}     &      \textbf{59.19}       &      \textbf{70.75}    &    77.30       &    \textbf{74.00}     \\
			&  M$_G$ &    69.89   &   71.30     &    59.97   &    82.19     &      78.75    &     \textbf{73.32}      &   69.96       &        74.85       &    \textbf{82.56}     &     \textbf{82.85}     &       56.20      &     66.37     &     75.33      &    72.58    \\ \hline
			\multirow{4}{*}{2}     & S &   \textbf{79.64}    &    61.08    &   77.57    &    77.86     &     74.10     &     \textbf{77.31}      &    79.24      &        74.96       &    \textbf{80.02}     &    79.60      &       \textbf{67.29}      &     65.63     &     75.26      &    74.58     \\
			& P  &   73.60    &    73.91    &   83.81    &    79.62     &     75.22     &     73.62      &     83.21     &       74.27        &    71.32     &     80.13     &      64.96       &     72.56     &      78.56     &   75.75      \\
			& M$_S$   &    79.01   &    \textbf{76.36}    &    82.34   &    \textbf{79.75}     &    \textbf{77.20}      &     73.32      &    84.39      &       74.14        &    76.16     &    \textbf{81.58}      &       67.27      &     \textbf{77.41}     &     \textbf{79.26}      &    \textbf{77.55}     \\
			& M$_G$   &   76.01    &    72.22    &    \textbf{83.82}   &     79.08    &     76.83     &     68.34      &   \textbf{84.62}       &        \textbf{75.07}       &    73.92     &    80.92      &       65.97      &     77.02     &     79.25      &    76.39   \\ \bottomrule
		\end{tabular}
	}
\end{center}
\end{table*}

\subsection{Quantitative Results}
\label{sec:Quantitative Results}
\subsubsection{UCF-Crime}
\label{sec:UCF-Crime}
We conduct our experiments on two action recognition data splits in the UCF-Crime dataset.
Even though UCF-Crime is the largest dataset for anomaly detection in videos, it is still too small to conduct meta-learning experiments.
We generated 13 subtasks for the experiment, where each subtask has a different novel subclass while the rest of 12 subclasses are employed to construct $\mathcal{D}_{base}$.    

Table~\ref{table:result} reports the AUC scores for all 13 subtasks, where we compare the results from the three scenarios described in Section~\ref{sec:Experimnetal Settings and Implementation Details}.
Since the statistics of each subclass are different from each other, it is not straightforward to identify the optimal model for a fair comparison.
Hence, we choose the following two different models for the evaluation of each subtask.
First, we sample 10 models for each subtask from the uniformly sampled meta-iterations, and select the best model for each subclass.
We call this model selection strategy ``sampling".
Second, to make the evaluation more strict, we choose a global model from the same meta-iteration to handle all subclasses, which is called ``global''.
For both splits, a meta-trained model shows better average performance than the others; the proposed  model improves accuracy in most of the subclasses while we observe accuracy drops by pretraining and meta-training in a few cases including \textit{Explosion} and \textit{Shoplifting} in the second split.
This is probably because their data distributions of these two subclasses are substantially different from the others and the prior knowledge is not helpful.
However, in the case of \textit{Stealing}, our approach outperforms the pretraining method by about 5\% margin, which indicates the proposed technique is effective even for the scenario that pretraining does not help.

\subsubsection{ShanghaiTech}
\label{sec:ShanghaiTech}
To validate the generalization capability of our model, we conduct an additional experiment on the ShanghaiTech dataset~\cite{liu2018ano_pred}.
As mentioned in Section~\ref{sec:Datasets}, it is not feasible to perform meta-training on this dataset due to its small size.
Since the examples in ShanghaiTech belong to abnormal events in campus life, which are unique compared to UCF-Crime, we consider the ShanghaiTech dataset as $\mathcal{D}_{novel}$ and use UCF-Crime as $\mathcal{D}_{base}$ to learn the prior knowledge.
All hyperparameters of the experiments for ShanghaiTech are identical to those for UCF-Crime.

\begin{table*}[t!]
	\caption{
		Quantitative results on the ShanghaiTech dataset.
		For P and M$_S$, UCF-Crime dataset is employed to train the model.
		Then, each model is fine-tuned with a train split of ShanghaiTech, and the final evaluation is conducted on the test split of ShanghaiTech. 
		The results show that the meta-initialized model adapts better to novel anomaly than the others, S and P.
	}
	\label{table:result_shanghai}
\begin{center}
	\scalebox{0.82}{
	\hspace{-0.3cm}
		\begin{tabular}{@{}M{3cm} |M{3cm} @{}}
			\toprule
		 	Algorithm & AUC (\%)\\ \hline
			S   &    79.53 \\ 
			P  &    79.34\\
			M$_S$   &   \textbf{84.70} \\  \bottomrule
		\end{tabular}
	}\vspace{-5mm}
\end{center}
\end{table*}

Table~\ref{table:result_shanghai} presents the results of the three training scenarios. 
The proposed strategy also outperforms the other training methods, which implies that the knowledge from one dataset learned by our meta-learning approach is transferable to other datasets; the proposed framework provides a promising initial model for localizing abnormalities in diverse situations.

\begin{figure}[!h]
	\centering
	\begin{subfigure}[t]{1\textwidth}
		\raisebox{-\height}{\includegraphics[width=\textwidth]{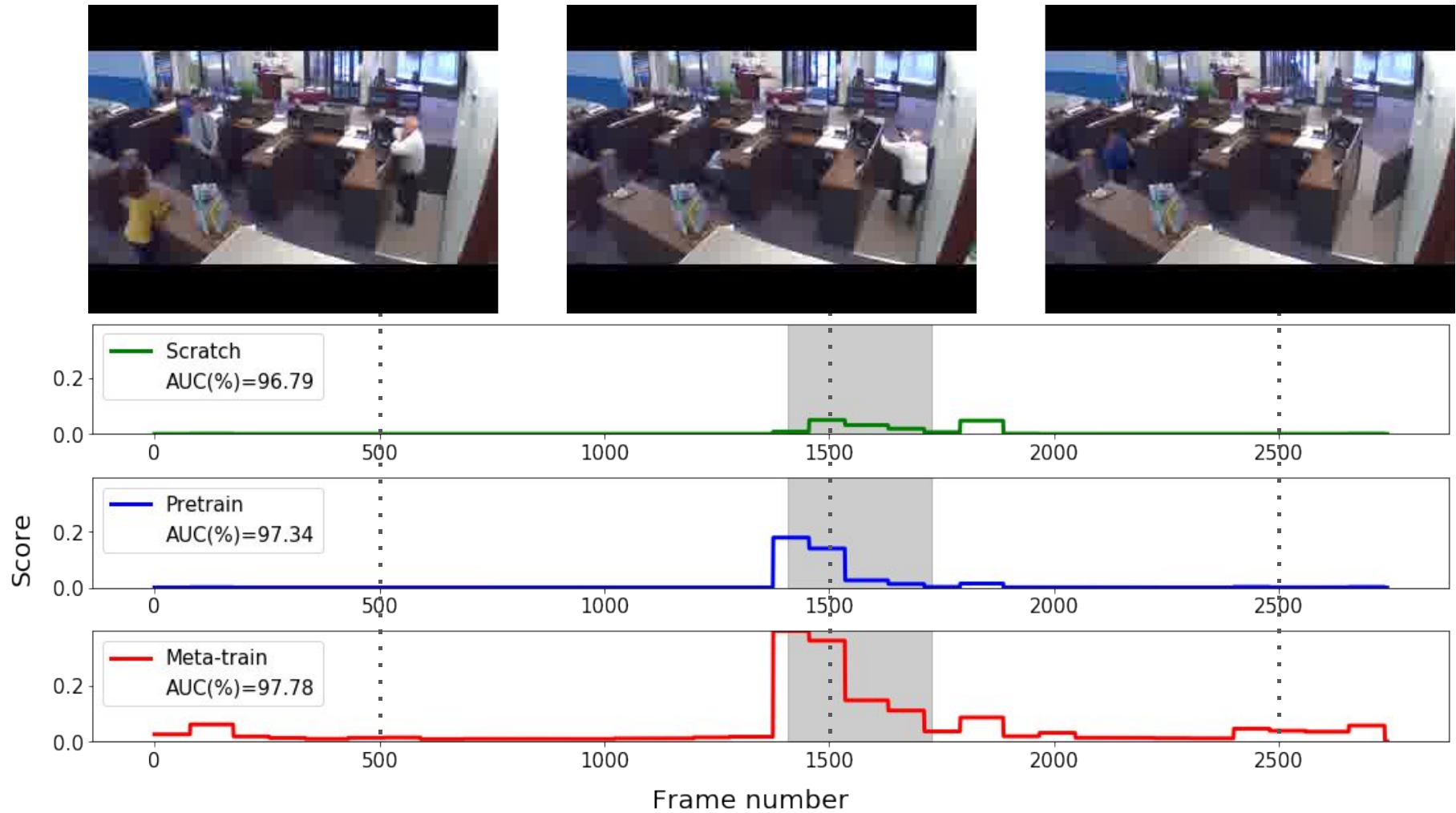}}
		\caption{ Shooting048 sequence in UCF-Crime dataset}
		\vspace{0.3cm}
	\end{subfigure}
	\begin{subfigure}[t]{1\textwidth}
		\raisebox{-\height}{\includegraphics[width=\textwidth]{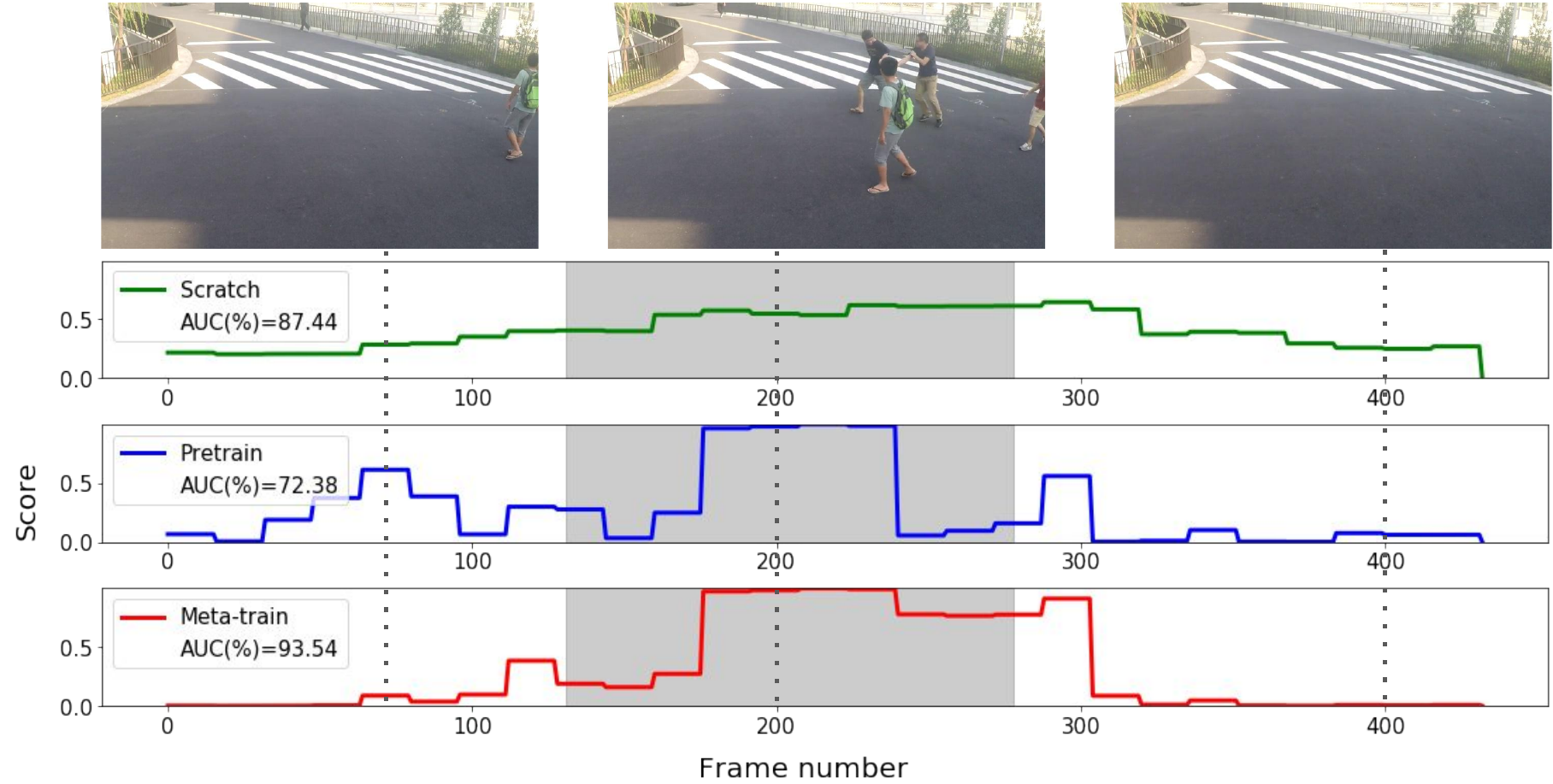}}
		\caption{03\_0059 sequence in ShanghaiTech dataset}
	\end{subfigure}
	
	\caption{Qualitative results from (a) the UCF-Crime and (b) the ShanghaiTech datasets. The scores of three different methods are presented together with the ground-truths represented by the shaded regions.}
	\label{fig:qual}
\end{figure}

\subsection{Qualitative results}
Fig.~\ref{fig:qual} demonstrates the qualitative results from three training scenarios on two test videos in (a) the UCF-Crime and (b) ShanghaiTech datasets.
The shaded regions in the graphs correspond to the ground-truth intervals of abnormal events.
The area under the ROC curve (AUROC or AUC) for each video and model is also reported in the graph.
Since the AUC metric is computed by the rank of scores, the performance of all the three methods looks comparable.
However, the scores given by the meta-trained model are more discriminative than the other two methods.
In other words, the models trained from scratch or pretrained models are prone to suffer from mis-detections and/or false alarms.
This observation implies that the proposed approach would be more robust than the others in more challenging examples.
%
We will enclose more sample results of anomaly detection with scores in the supplementary document.

\begin{figure}[h!]
	\centering
	\begin{subfigure}[t]{0.325\textwidth}
		\raisebox{-\height}{\includegraphics[width=\textwidth]{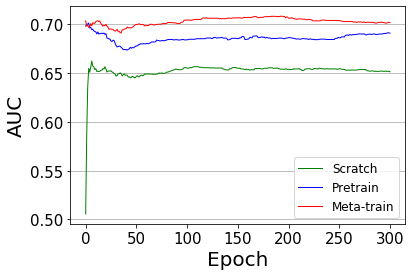}}
		\caption{Abuse}
	\end{subfigure}
	\vspace{0.1cm}
	\begin{subfigure}[t]{0.325\textwidth}
		\raisebox{-\height}{\includegraphics[width=\textwidth]{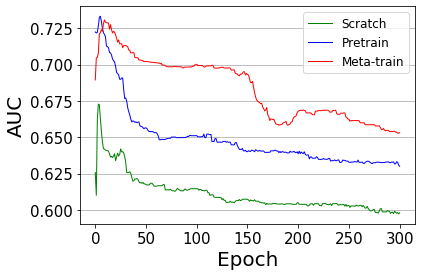}}
		\caption{Explosion}
	\end{subfigure}
	\vspace{0.1cm}
	\begin{subfigure}[t]{0.325\textwidth}
		\raisebox{-\height}{\includegraphics[width=\textwidth]{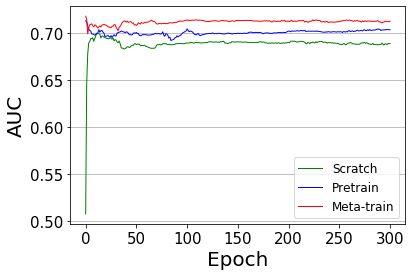}}
		\caption{Fighting}
	\end{subfigure}
	\vspace{0.1cm}
	\begin{subfigure}[t]{0.325\textwidth}
		\raisebox{-\height}{\includegraphics[width=\textwidth]{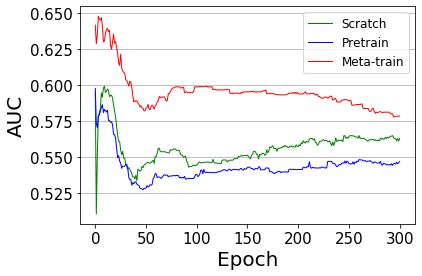}}
		\caption{Shoplifting}
	\end{subfigure}
	\begin{subfigure}[t]{0.325\textwidth}
		\raisebox{-\height}{\includegraphics[width=\textwidth]{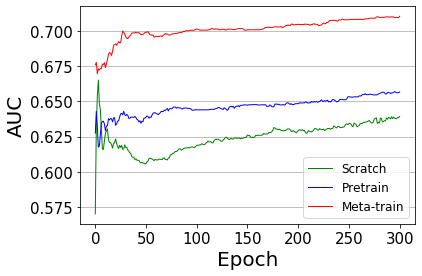}}
		\caption{Stealing}
	\end{subfigure}
	\begin{subfigure}[t]{0.325\textwidth}
		\raisebox{-\height}{\includegraphics[width=\textwidth]{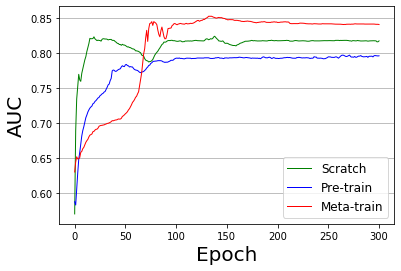}}
		\caption{S-Tech}
	\end{subfigure}

	\caption{Fine-tuning curves of individual subclasses for three training scenarios}
	\label{fig:learning_curve}
\end{figure}

\subsection{Further Analysis}
\label{sec:Further Analysis}
To analyze whether the meta-learning scheme has a real impact on adapting to novel abnormalities, we plot performance curves during fine-tuning at the meta-testing stage with the UCF-Crime and ShanghaiTech datasets and present the results in Fig.~\ref{fig:learning_curve}.
The followings are the lessons from the observation of the fine-tuning curves.

First, the anomaly detection benefits from prior knowledge.
In most cases, both of the pretrained model and the meta-trained model demonstrate better performance at the initial epoch than the scratch case.
From this observation, we conclude that the model has the utility to detect the novel types of abnormal events even without their direct prior knowledge.
Second, the performance of a model is sometimes degraded during the fine-tuning procedure.
Fig.~\ref{fig:learning_curve}(b), and (d) illustrates that the learning curves go downward with iterations.
For these cases, we conclude that there are data samples with significant noise or large intra-class variations; it is challenging for a model to detect anomalies with weak supervision only.
This stems from the inherent weakness of the detector trained based only on weak supervision.  
Due to space limitations, the learning curves for the rest of the subclasses are included in the supplementary document.


\section{Conclusion}
\label{sec:conclusion}
We presented a weakly supervised learning-to-adapt formulation of anomaly detection in videos, which alleviates the limitation of existing methods in the generalization to diverse unseen data samples.
To this end, we proposed a learning strategy to adapt to unseen types of abnormal events effectively by taking advantage of meta-learning.
We meta-train the model by constructing episodes that are well-aligned with anomaly detection.
Our experimental results from challenging UCF-Crime and ShanghaiTech demonstrate that the models given by the proposed technique learn to adapt to new types of abnormal videos successfully and verify the efficacy of meta-learning in adaptation quality compared to the pretrained models.
In addition, we pointed out the limitation of UCF-Crime dataset in terms of annotation completeness and data imbalance, and supplement temporal annotations of abnormal activities for the videos which do not have such ground-truths.

\subsubsection{Acknowledgments}
This work was partly supported by Vision AI Product Center of Excellence in T3K of SK telecom and Institute for Information \& Communications Technology Promotion (IITP) grant funded by the Korea government (MSIT) [2017-0-01779, 2017-0-01780].

\clearpage

\bibliographystyle{splncs}
\bibliography{egbib}

\end{document}


\pagestyle{headings}
\mainmatter
\def\ACCVSubNumber{479}  

\title{Supplementary Material for \\
	Learning to Adapt to Unseen Abnormal
	Activities under Weak Supervision} 

\titlerunning{Learning to Adapt to Unseen Abnormal Activities}
%
%
\author{Jaeyoo Park\thanks{These authors contributed equally.} \and 
Junha Kim\textsuperscript{*} \and 
Bohyung Han}

\authorrunning{J. Park, J. Kim, and B. Han}
%
\institute{ECE \& ASRI, Seoul National University, Korea \\
\email{\{bellos1203,junha.kim,bhhan\}@snu.ac.kr}}
\maketitle





\section{Supplementary}

\subsection{Multiple classes for meta-test}
To verify that our algorithm works well for the meta-test task consisting of multiple abnormal subclasses, we conduct an additional experiment. 
We randomly split 13 subclasses of the UCF-Crime dataset into 7 for meta-train and 6 for meta-test, and vice versa (6 for meta-train and 7 for meta-test).
We aggregate the results and report the average performance.
We conduct the experiments for two randomly sampled subclass splits. 
Table.~\ref{table:results} describes the results.
The results show that meta-learning performance is still better than the others, even with multiple meta-test subclasses, and the overall results are similar to Table 2. in the main paper.

\begin{table}[t]
	\caption{
		Quantitative results where meta-test task contains multiple subclasses.
		The different split number denotes the different split of 7 and 6 subclasses while, in Table 2. of the main paper, it indicates different split of data samples.
		In this experiment, we use the first data sample split to construct the support set and query set of the meta-test.
	}
	\label{table:results}
\begin{center}
	\scalebox{0.95}{
	\renewcommand{\arraystretch}{1.1}
	\setlength\tabcolsep{10pt}
	\begin{tabular}{c|c|c}
		\toprule
		Split          & Algorithm & AUC (\%)  \\
    		\hline
		\multirow{3}{*}{1}        & S   & 71.30                    \\
		 & P & 70.75	\\
		 & M$_S$ & 72.69	\\ \hline
		\multirow{3}{*}{2}        & S   & 70.32                    \\
		 & P & 70.36	\\
		 & M$_S$ & 72.90 \\
		\bottomrule                
	\end{tabular}
	}
\end{center}
\end{table}

\subsection{Additional Qualitative Results}
\begin{figure}[h]
	\centering
	\begin{subfigure}[t]{1\textwidth}
		\raisebox{-\height}{\includegraphics[width=\textwidth]{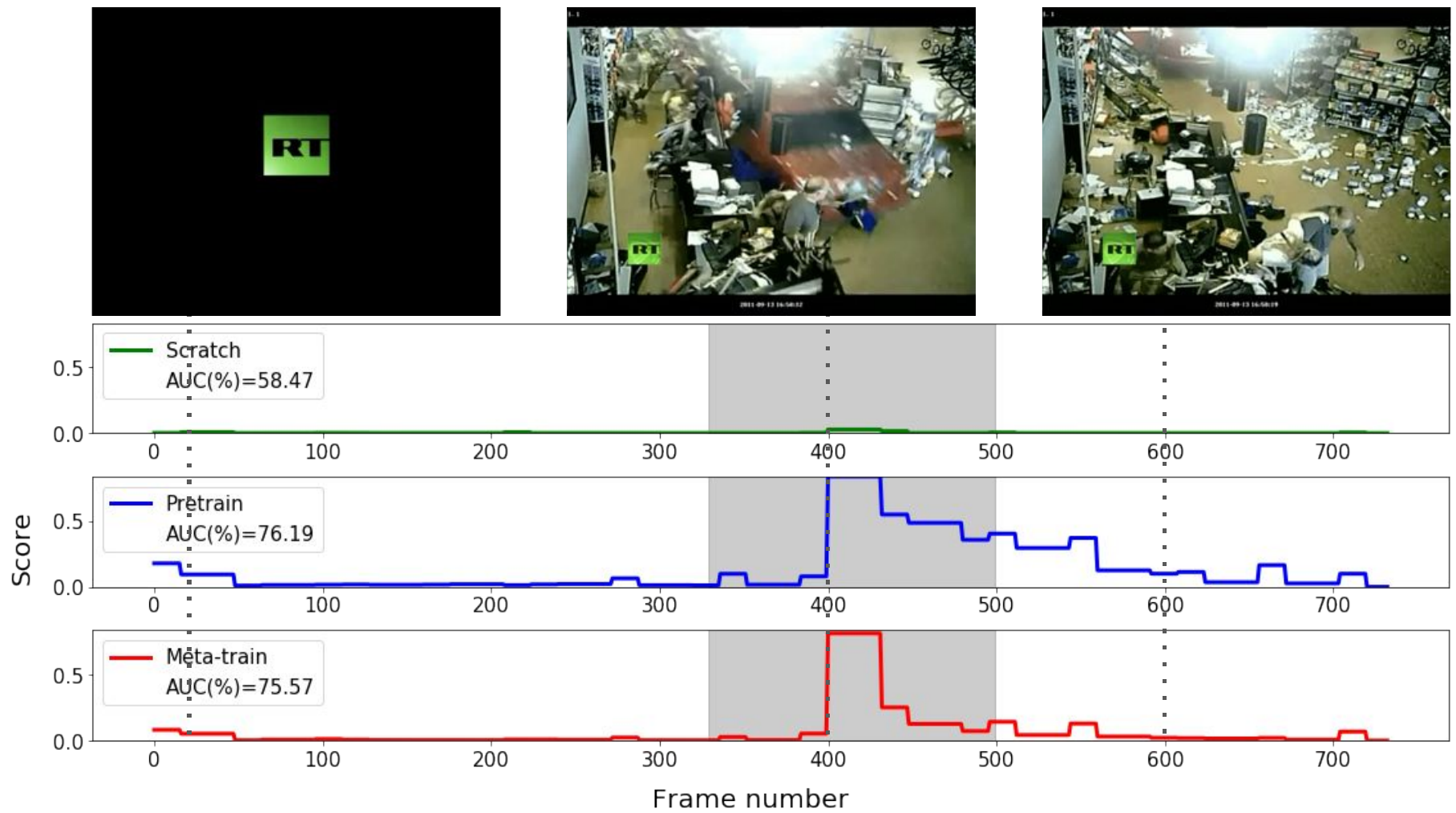}}
		\caption{RoadAccidents049 sequence in UCF-Crime dataset}
	\end{subfigure}
	\begin{subfigure}[t]{1\textwidth}
		\raisebox{-\height}{\includegraphics[width=\textwidth]{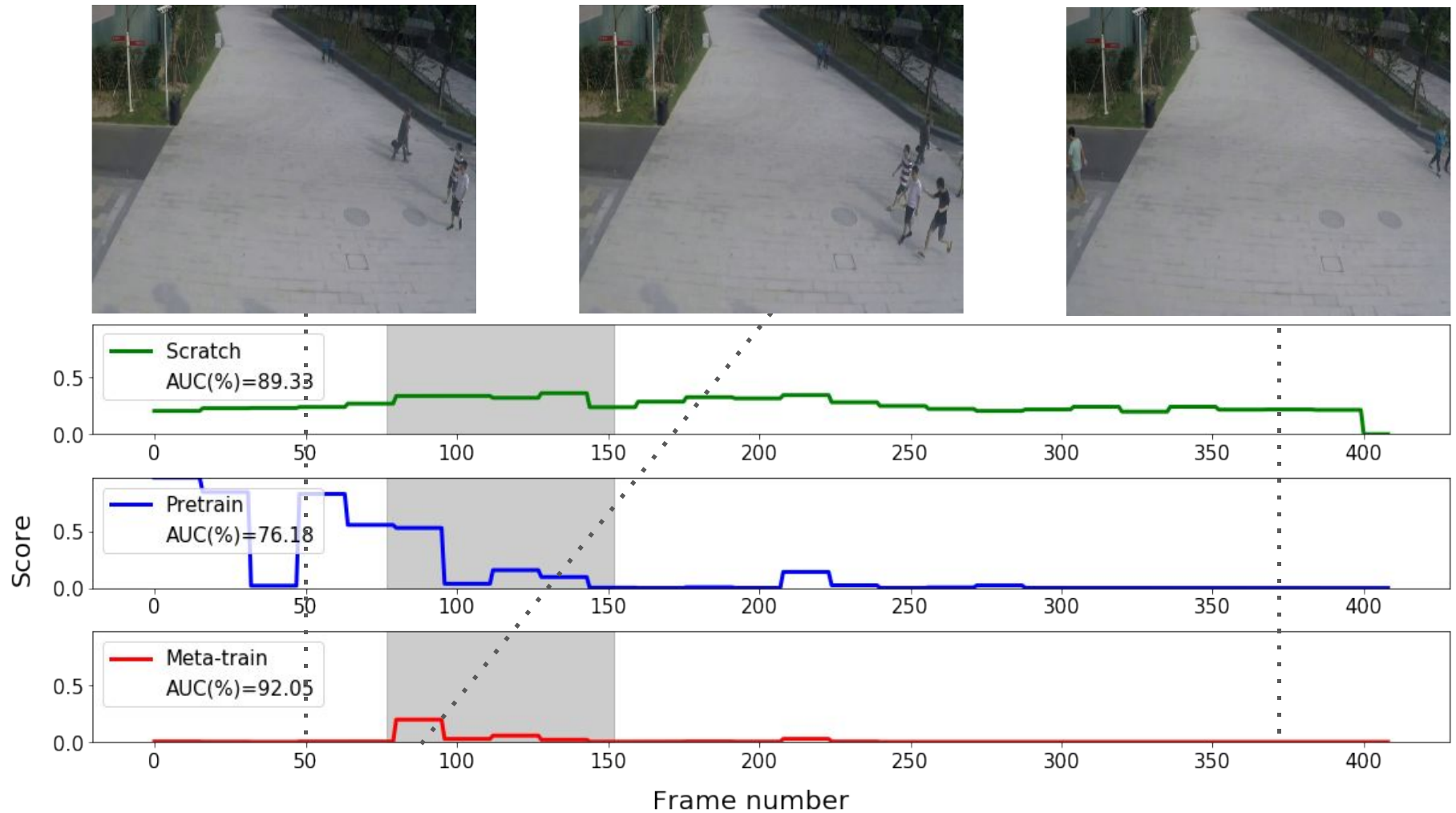}}
		\caption{01\_0027 sequence in ShanghaiTech dataset}
	\end{subfigure}

	\caption{Qualitative results from UCF-Crime and ShanghaiTech datasets. The scores of three different methods are presented together with the ground-truth represented by the shaded regions correspond to the ground-truth.}
	\label{fig:qual}
\end{figure}

Fig.~\ref{fig:qual} illustrates the additional qualitative results from three methods on three test videos in the UCF-Crime and ShaghaiTech.
Following Fig. 3 in the main paper, the shaded regions in the graphs correspond to the ground-truth intervals of abnormal events. 
The area under the ROC curve (AUROC or AUC) for each video and model is also reported in the graph. 
The scores given by the meta-trained model are much more discriminative than the other two methods.
In (a), the scratch model does not capture the abnormalities well compared to the other two models, which take advantage of prior knowledge.
Additionally, we observed that the pretrained and meta-trained models are activated by the frames containing the logo with the background colored in black, while the models still distinguish well those frames from the ones containing the ground truth abnormalities.
In (b), the figure illustrates that the models trained from scratch or pretrained models are prone to suffer from mis-detections and/or false alarms while the meta-trained models maintain a better balance between positive and negative scores.
We attached four sample videos, including the main paper's videos, with the three methods' scores.

\subsection{Fine-tuning Curve Results for All Classes}
\begin{figure}[h]
	\centering
	\begin{subfigure}[t]{0.325\textwidth}
		\raisebox{-\height}{\includegraphics[width=\textwidth]{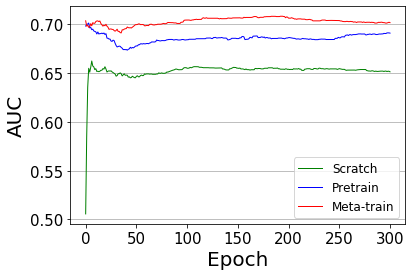}}
		\caption{Abuse}
	\end{subfigure}
	\begin{subfigure}[t]{0.325\textwidth}
		\raisebox{-\height}{\includegraphics[width=\textwidth]{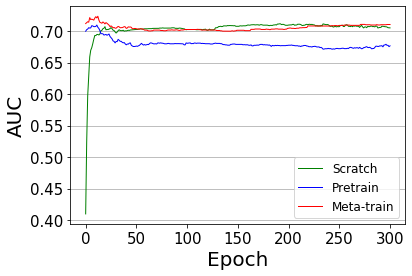}}
		\caption{Arrest}
	\end{subfigure}
	\begin{subfigure}[t]{0.325\textwidth}
		\raisebox{-\height}{\includegraphics[width=\textwidth]{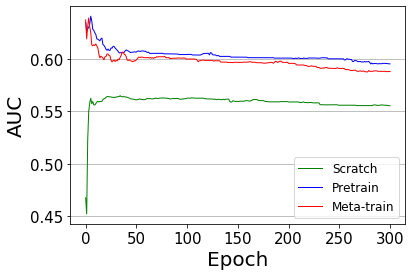}}
		\caption{Arson}
	\end{subfigure}
	\begin{subfigure}[t]{0.325\textwidth}
		\raisebox{-\height}{\includegraphics[width=\textwidth]{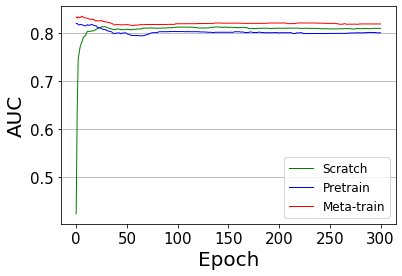}}
		\caption{Assault}
	\end{subfigure}
	\begin{subfigure}[t]{0.325\textwidth}
		\raisebox{-\height}{\includegraphics[width=\textwidth]{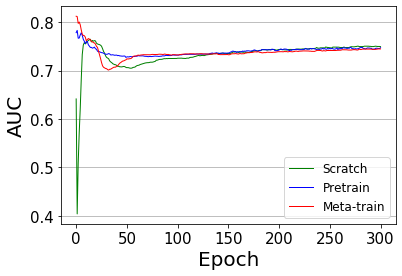}}
		\caption{Burglary}
	\end{subfigure}
	\begin{subfigure}[t]{0.325\textwidth}
		\raisebox{-\height}{\includegraphics[width=\textwidth]{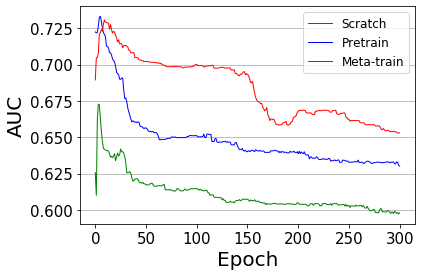}}
		\caption{Explosion}
	\end{subfigure}
	\begin{subfigure}[t]{0.325\textwidth}
		\raisebox{-\height}{\includegraphics[width=\textwidth]{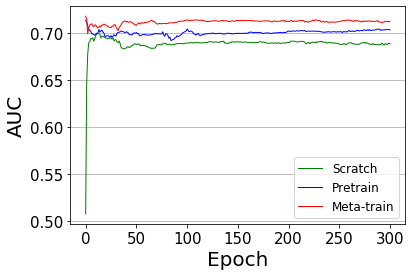}}
		\caption{Fighting}
	\end{subfigure}
	\begin{subfigure}[t]{0.325\textwidth}
		\raisebox{-\height}{\includegraphics[width=\textwidth]{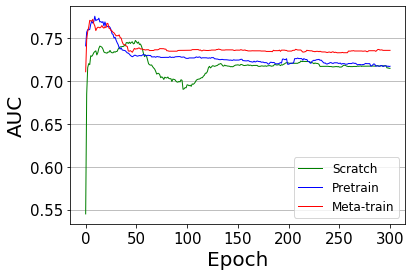}}
		\caption{RoadAcc}
	\end{subfigure}
	\begin{subfigure}[t]{0.325\textwidth}
		\raisebox{-\height}{\includegraphics[width=\textwidth]{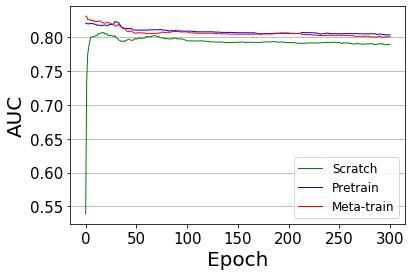}}
		\caption{Robbery}
	\end{subfigure}
	\begin{subfigure}[t]{0.325\textwidth}
		\raisebox{-\height}{\includegraphics[width=\textwidth]{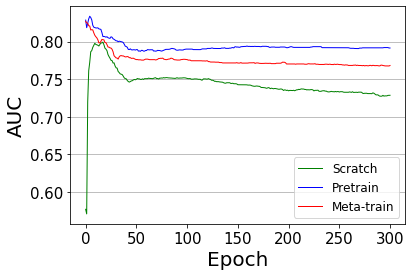}}
		\caption{Shooting}
	\end{subfigure}
	\begin{subfigure}[t]{0.325\textwidth}
		\raisebox{-\height}{\includegraphics[width=\textwidth]{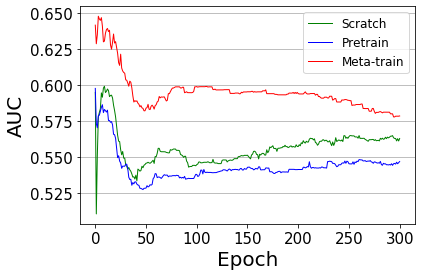}}
		\caption{Shoplifting}
	\end{subfigure}
	\begin{subfigure}[t]{0.325\textwidth}
		\raisebox{-\height}{\includegraphics[width=\textwidth]{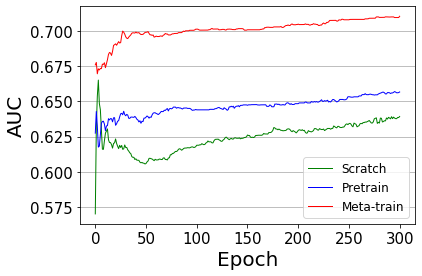}}
		\caption{Stealing}
	\end{subfigure}
	\begin{subfigure}[t]{0.325\textwidth}
		\raisebox{-\height}{\includegraphics[width=\textwidth]{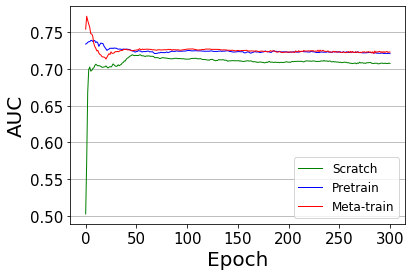}}
		\caption{Vandalism}
	\end{subfigure}
	\begin{subfigure}[t]{0.325\textwidth}
		\raisebox{-\height}{\includegraphics[width=\textwidth]{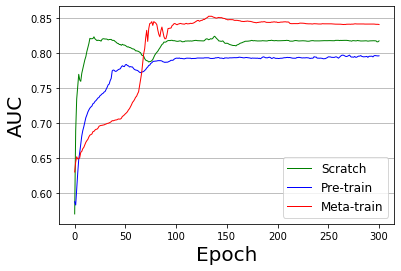}}
		\caption{S-Tech}
	\end{subfigure}

	\caption{Comparison of subclass-wise fine-tuning curve for each scenario}
	\label{fig:learning_curve}
\end{figure}

We attach additional fine-tuning graphs for each scenario in this supplementary material, including the graphs for the subclasses omitted in the main paper due to space.
The results in Fig.~\ref{fig:learning_curve} show the final performance curves on the test dataset during the fine-tuning phase. 
Although there are some cases where the meta-trained model is not the best, the performance of meta-trained cases shows better than or comparable to that of the other cases, i.e., the scratch and pre-trained model.
In failure cases like (c) or (j), the margin between the best and meta-trained models is quite negligible, given the margin of successful cases.
In addition, the performance of the meta-trained model is still better than the scratch model, even in the failure cases.
Please refer to Section 4.6 of the main paper for the details.

